\documentclass[journal]{IEEEtran} 
\usepackage[T1]{fontenc}
\usepackage[latin9]{inputenc}
\usepackage{color}
\usepackage[dvipsnames]{xcolor}
\usepackage{float}
\usepackage{textcomp}
\usepackage{mathrsfs}
\usepackage{mathtools}
\usepackage{amsmath}

\usepackage{amsthm}
\usepackage{amssymb}
\usepackage{stmaryrd}
\usepackage{stackrel}
\usepackage{graphicx}
\usepackage{wasysym}
\usepackage{url}
\usepackage[font=small,labelfont=bf]{caption}
\usepackage{comment}
\usepackage{makecell}
\usepackage{multirow}
\makeatletter

\floatstyle{ruled}
\newfloat{algorithm}{tbp}{loa}
\providecommand{\algorithmname}{Algorithm}
\floatname{algorithm}{\protect\algorithmname}

\theoremstyle{plain}
\newtheorem{thm}{\protect\theoremname}
\theoremstyle{definition}
\newtheorem{defn}{\protect\definitionname}
\theoremstyle{definition}

\theoremstyle{plain}

\theoremstyle{definition}
\newtheorem{example}{\protect\examplename}
\theoremstyle{remark}

\theoremstyle{plain}

\providecommand{\lemmaname}{Lemma}
\usepackage{cite}
\usepackage{algpseudocode}
\usepackage{setspace}
\IEEEoverridecommandlockouts
\makeatother

\usepackage{babel}
\providecommand{\definitionname}{Definition}
\providecommand{\examplename}{Example}
\providecommand{\problemname}{Problem}
\providecommand{\theoremname}{Theorem}

\usepackage{color}

\begin{document}
% \pagenumbering{gobble}

 \title{Temporal Logic Guided Motion Primitives for Complex Manipulation Tasks
with User Preferences \thanks{H. Wang, H. He, W. Shang and Z. Kan (Corresponding Author) are with
the Department of Automation at the University of Science and Technology
of China, Hefei, Anhui, China, 230026. }}
\author{Hao Wang, Haoyuan He, Weiwei Shang, and Zhen Kan}
\maketitle
\begin{abstract}
Dynamic movement primitives (DMPs) are a flexible trajectory learning
scheme widely used in motion generation of robotic systems. However,
existing DMP-based methods mainly focus on simple go-to-goal tasks.
Motivated to handle tasks beyond point-to-point motion planning, this
work presents temporal logic guided optimization of motion primitives,
namely $\textrm{P\textrm{I}}^{\textrm{BB-TL}}$ algorithm, for complex
manipulation tasks with user preferences. In particular, weighted
truncated linear temporal logic (wTLTL) is incorporated in the $\textrm{P\textrm{I}}^{\textrm{BB-TL}}$
algorithm, which not only enables the encoding of complex tasks that
involve a sequence of logically organized action plans with user preferences,
but also provides a convenient and efficient means to design the cost
function. The black-box optimization is then adapted to identify optimal
shape parameters of DMPs to enable motion planning of robotic systems.
The effectiveness of the $\textrm{P\textrm{I}}^{\textrm{BB-TL}}$
algorithm is demonstrated via simulation and experiment.%
\begin{comment}
\begin{abstract}
\global\long\def\Dist{\operatorname{Dist}}%
\global\long\def\Inf{\operatorname{Inf}}%
\global\long\def\Sense{\operatorname{Sense}}%
\global\long\def\Eval{\operatorname{Eval}}%
\global\long\def\Info{\operatorname{Info}}%
\global\long\def\ResetRabin{\operatorname{ResetRabin}}%
\global\long\def\Post{\operatorname{Post}}%
\global\long\def\Acc{\operatorname{Acc}}%
\end{abstract}

\end{comment}

\global\long\def\Sign{\operatorname{Sign}}%
\end{abstract}

\section{INTRODUCTION}

By operating beyond structured environments, robots are moving towards
applications in complex and unstructured environments, including offices,
hospitals, homes \cite{Chatzilygeroudis2020}. These applications
often require robots to be able to autonomously learn, plan, and execute
a variety of challenging manipulations. As an enabling technique,
dynamic movement primitives (DMPs) have emerged as a flexible trajectory
learning scheme \cite{Schaal2006,Bottasso2008,Dutta2018} for manipulators
and mobile robots, such as biped walking of humanoid robots \cite{Rakovic2014},
collaborative manipulation of clothes \cite{Colome2018}, bimanual
tasks \cite{Gams2014}. While it is powerful, conventional DMPs are
mainly limited to simple go-to-goal tasks. Another challenge that
receives little attention in prior DMP-based approaches is the problem
of capturing user preferences during motion and trajectory generation.
Therefore, this work is particularly motivated to improve DMP-based
motion generation skills for robotic systems to perform manipulations
that consist of a sequence of logically organized actions with user
preferences. 

\textbf{Related works: }DMPs are encoded as a combination of simple
linear dynamical systems with nonlinear components to acquire smooth
movements of arbitrary shape. One common objective is to obtain the
optimal parameters of nonlinear components. As discussed in \cite{Ijspeert2013},
the trajectory generated by DMPs has good scaling properties with
respect to the initial/end positions, the parameter linearity, the
rescaling robustness, and the continuity. For these reasons, DMPs
have been widely used with policy search reinforcement learning to
identify a best policy (i.e., the optimal DMP parameters). Example
policy search methods include gradient based approaches \cite{Peters2008},
expectation maximization based approaches \cite{Peters2006}, and
information-theoretic approaches \cite{Daniel2016}. The policy improvement
with path integrals algorithm $\textrm{P\textrm{I}}^{\textrm{2}}$
was derived from the first principles of stochastic optimal control
\cite{Stulp2012,Theodorou2010}. Different with gradient-based reinforcement
learning algorithms, the $\textrm{P\textrm{I}}^{\textrm{2}}$ algorithm
avoids the curse of dimensionality and gradient estimation by using
parameterized policies with probability-weighted averaging. However,
the motion primitive parameters for each dimension must be optimized
individually when using $\textrm{P\textrm{I}}^{\textrm{2}}$ algorithm,
which increases the computational complexity and reduces the parameter
convergence rate. A stochastic optimization algorithm, namely Policy
Improvement with Black-Box Optimization $\textrm{P\textrm{I}}^{\textrm{BB}}$,
was then developed in \cite{Stulp2018}. As a special case of $\textrm{P\textrm{I}}^{\textrm{2}}$,
$\textrm{P\textrm{I}}^{\textrm{BB}}$ uses the same method for exploration
and parameter updating, but differs in using black-box optimization
rather than reinforcement learning for policy improvement. Since $\textrm{P\textrm{I}}^{\textrm{BB}}$
is based on covariance matrix adaptation through weighted averaging,
it can optimize the parameters of motion primitives with multiple
dimensions simultaneously for improved efficiency and convergence
rate. Other representative works that exploit DMP for motion generation
include \cite{Stulp2012,Gams2014,Calinon2016}. However, limited to
the form of the expect cost function, neither $\textrm{P\textrm{I}}^{\textrm{BB}}$
nor $\textrm{P\textrm{I}}^{\textrm{2}}$ can handle parameter optimization
of motion primitives for manipulation tasks with complex logic and
temporal constraints. 

Temporal logic, as a formal language, is capable of describing a wide
range of complex tasks in a succinct and human-interpretable form,
and thus has been increasingly used in the motion planning of robotic
systems \cite{Baier2008,Kloetzer2015,Cai2020,Cai2021a,Cai2021,Bozkurt2020}.
Signal temporal logic (STL) is defined over continuous signals and
its quantitative semantics, known as robustness, can measure the degree
of satisfaction or violation of the desired task specification \cite{Maler2004}.
To maximize the robustness of STL, the synthesis problem is often
cast as optimization problems and then solved using heuristics, mixed-integer
programming or gradient methods \cite{Saha2016,Raman2014,Belta2019,Mehdipour2021,Puranic2021}.
However, STL formulas specify subtasks with explicit time bounds,
while many manipulation tasks only require the subtasks to be performed
in a desired sequence (e.g., open the fridge door, take the milk out,
and close the fridge door). Manually assigning time bounds for subtasks
might lead to the failure of finding desired policy due to unexpected
environmental events. Other possible formal languages, such as BLTL
\cite{Latvala2004} and $\textrm{LT\ensuremath{\textrm{L}_{f}}}$
\cite{DeGiacomo2013}, either require time bounds similar to STL or
does not come with quantitative semantics. In contrast to the aforementioned
methods, truncated linear temporal logic (TLTL) is a predicate temporal
logic without time bounds, which is defined over finite-time trajectories
of robot\textquoteright s states and provides a unifying and interpretable
way to specify tasks \cite{Li2017a}. The quantitative semantics of
TLTL, also referred to as robustness, can be used to design interpretable
reward or cost for motion planning of robotic systems. In \cite{Li2019},
TLTL was successfully used to specify a robotic cooking task. Despite
its recent success, TLTL is mainly used for high-level motion planning
and few effort has been devoted to extending TLTL with general trajectory
generation approaches, such as DMP based methods. 

\textbf{Contributions:} In this work, we consider motion generation
for a robotic system to perform logically and temporally structured
manipulations with user preferences. Since conventional DMP based
approaches (e.g., $\textrm{P\textrm{I}}^{\textrm{2}}$ or $\textrm{P\textrm{I}}^{\textrm{BB}}$)
suffer from handcrafted cost function and are mainly used for simple
point-to-point tasks, the first contribution is to develop the $\textrm{P\textrm{I}}^{\textrm{BB-TL}}$
algorithm by extending the state-of-the-art motion generation $\textrm{P\textrm{I}}^{\textrm{BB}}$
algorithm with TLTL, so that it is able to learn optimal shape parameters
of motion primitives for tasks with logic and temporal constraints.
Compared with most existing works, the use of TLTL not only enables
the encoding of complex tasks that involve a sequence of logically
organized action plans, but also provides a convenient and effective
means to design the cost function. Close to our work, LTL specifications
was also incorporated in the learning of DMPs in \cite{Innes2020}.
However, the loss function designed in \cite{Innes2020} is limited
in evaluating whether or not a given LTL specification is satisfied
and the log-sum-exponential approximation is over-approximated. In
contrast, the weighted TLTL robustness in this work is sound that
can not only qualitatively evaluate the satisfaction of LTL specifications,
but also quantitively determine its satisfaction degree. Specifically,
the cost function is designed based on the TLTL robustness and the
smooth approximations in $\textrm{P\textrm{I}}^{\textrm{BB}}$ algorithm
to optimize the shape parameters of DMPs, ensuring that the generated
trajectory of DMPs satisfies complex tasks specified by TLTL constraints.
\begin{comment}
Since traditional robustness semantics only considers satisfaction
of a formula at the most extreme sub-formula, hindering the optimization
to find a more robust solution, in this work, based on smooth approximations
\cite{Gilpin2020}, TLTL robustness is refined by accumulating/averaging
the robustness of all the sub-formulae. It should be noted that $\textrm{P\textrm{I}}^{\textrm{2}}$SEQ,
an extend $\textrm{P\textrm{I}}^{\textrm{2}}$ algorithm, is also
applicable to the trajectory generation for a sequence of tasks \cite{Stulp2012};
however, its high computation complexity might limit its practical
use, e.g., given a $n$-goal-reaching task, $n$ DMPs need to be trained
using $\textrm{P\textrm{I}}^{\textrm{2}}$SEQ while only one is needed
using $\textrm{P\textrm{I}}^{\textrm{BB-TL}}$ to generate a satisfactory
trajectory.
\end{comment}
{} Another contribution is to take into account user preferences in
motion generation. Inspired by \cite{Mehdipour2021}, we further extend
TLTL to weighted TLTL (wTLTL) to capture sub-task with different importance
or priorities. Incorporating DMPs with wTLTL ensures that the generated
trajectory satisfies the given manipulation task with user specified
preference. %
\begin{comment}
Therefore, these contributions of $\textrm{P\textrm{I}}^{\textrm{BB-TL}}$
algorithm advance path integral policy improvement beyond learning
only shape parameters with single motion primitives.
\end{comment}
{} The effectiveness of $\textrm{P\textrm{I}}^{\textrm{BB-TL}}$ is
demonstrated via simulation and experimental results. 

\section{PRELIMINARIES}

In this work, since DMPs and weighted TLTL form the basic building
blocks of $\textrm{P\textrm{I}}^{\textrm{BB-TL}}$ algorithm, they
are briefly introduced in this section.

\subsection{Dynamic Movement Primitives\label{subsec:DMPs}}

Dynamic movement primitives are a flexible representation of robot
trajectories \cite{Schaal2006}, which can be expressed as
\begin{subequations}
\label{eq:DMP}
\begin{align}
\frac{1}{\tau}\dot{z}_{t} & =\alpha_{z}\left(\beta_{z}\left(g-y_{t}\right)-z_{t}\right)+f_{t},\label{eq:Transf.1}\\
\frac{1}{\tau}\dot{y}_{t} & =z_{t},\label{eq:Transf.2}\\
f_{t} & =\varsigma_{t}^{T}\mathbf{\mathbf{\theta}},\label{eq:Transf.3}\\
\frac{1}{\tau}\dot{s}_{t} & =-\alpha_{s}s_{t},\label{eq:Transf.4}
\end{align}
\end{subequations}
 where $y_{t}\in\mathbb{R}$ and $\dot{y}_{t}\in\mathbb{R}$ represent
the position and velocity, respectively, $z_{t}\in\mathbb{R}$ and
$s_{t}\in\mathbb{R}$ are internal states, $\alpha_{z}$, $\beta_{z}$,
$\tau$ and $\alpha_{s}$ are positive scale factors, $\mathbf{\theta\in\mathbb{R}}^{L}$
is the parameter vector, and $g\in\mathbb{R}$ is the goal position.
The nonlinear function $f_{t}$ allows the generation of arbitrary
complex movements, which consists of basis functions $\varsigma_{t}\in\mathbb{R}^{L}$
represented by a piecewise linear function approximator with weighted
Gaussian kernels $\varpi\in\mathbb{R}^{L}$ as 
\begin{align}
\left[\varsigma_{t}\right]_{i} & =\frac{\varpi_{i}\left(s_{t}\right)\cdot s_{t}}{\sum_{l=1}^{L}\varpi_{l}\left(s_{t}\right)}\left(g-x_{0}\right),\label{eq:Basis function}\\
\varpi_{i} & =\exp\left(-\frac{1}{2\sigma_{i}}\left(s_{t}-c_{i}\right)^{2}\right),\label{eq:Gassian kernel}
\end{align}
where $\left[\varsigma_{t}\right]_{i}$ denotes the $i$th entry of
$\varsigma_{t}$, and $\sigma_{i}$ and $c_{i}$ represent the variance
and mean, respectively. The core idea behind DMPs is to perturb the
term of $\alpha_{z}\left(\beta_{z}\left(g-y_{t}\right)-z_{t}\right)$
by a nonlinear $\varsigma_{t}^{T}\mathbf{\mathbf{\theta}}$ to acquire
smooth movements of arbitrary shape. Although only a 1-D system is
represented in (\ref{eq:DMP}), multi-dimensional DMPs can be represented
by coupling several dynamical systems using shared phase variable
$s_{t}$, where each dimension has its own goal $g$ and shape parameters
$\theta$. Reinforcement learning or black-box optimizations can then
be used to obtain the optimal shape parameters $\theta$.

\subsection{Weighted Truncated Linear Temporal Logic \label{subsec:Truncated-Linear-Temporal}}

Truncated linear temporal logic was introduced in \cite{Li2017a},
which is able to incorporate domain knowledge and various constraints
to describe complex robotic tasks. In \cite{Mehdipour2021}, weighted
signal temporal logic was developed to model user preferences. Inspired
by the works of \cite{Mehdipour2021} and \cite{Li2017a}, we extend
traditional TLTL to weighted TLTL (wTLTL) in this work to specify
complex robotic missions with weights modeling relative importance
and priority among temporal logic constraints (e.g., user preferences
over sub-tasks). 

The syntax of wTLTL is defined as
\begin{equation}
\begin{aligned}\varphi:= & \top\mid f\left(y_{t}\right)<c\mid\lnot\phi\mid\land^{w}\phi_{i}\mid\lor^{w}\psi_{i}\mid\\
 & \lozenge\phi\mid\Square\phi\mid\phi\mathcal{U}\psi\mid\phi\mathcal{T}\psi\mid\phi\Rightarrow\psi,
\end{aligned}
\end{equation}
where $\top$ is the boolean constant true, $f\left(y_{t}\right)<c$
is a predicate where $f\colon\mathbb{R^{\mathit{n}}\rightarrow\mathbb{R}}$
maps a system state $y_{t}$ to a constant, $\lnot$ (negation), $\lor$
(disjunction) and $\land$ (conjunction) are standard Boolean operators,
$\lozenge$ (eventually), $\boxempty$ (always), $\mathcal{U}$ (until),
$\mathcal{T}$ (then), and $\Rightarrow$ (implication) are temporal
operators. %
\begin{comment}
Due to the consideration of continuous time execution of robot motion,
the \textquotedblleft next\textquotedblright{} operator $\varbigcirc$
is no longer meaningful. For instance, consider a formula $\phi_{A}\varbigcirc\phi_{B}$
that requires the robot to visit region $B$ after visiting region
$A$, which is not practically feasible in continuous time execution
since it requires instantaneous motion. Therefore, in this work we
restrict our attention to TLTL specifications that exclude the next
operator. 
\end{comment}

Given $N$ conjunctions and disjunctions, the positive weight vector
is denoted by $w\in\mathbb{R}_{>0}^{N}$, where the $i$th entry $w_{i}$
associated with conjunction or disjunction indicates the corresponding
relative importance of obligatory specifications or priorities of
alternatives. Importance allows the trade-off of all specifications
that need to be satisfied (i.e., conjunctions), while priorities allows
the trade-off of the acceptance of alternative specifications (i.e.,
disjunctions). A higher value of $w_{i}$ corresponds to a higher
importance or priority. In the following sections, when the weight
vector of a Boolean operator ($\lor$ or $\land$) is a unit vector,
i.e. $w_{i}=1,$ $\forall i\in\left\{ 1,\ldots,N\right\} $, we omit
the $w$ in the wTLTL formula. Note that the traditional TLTL in \cite{Li2017a}
can be considered as a special case of wTLTL where all $w$ are unit
vectors.

The semantics of wTLTL formulas is defined over finite trajectories
of system states (e.g., a trajectory generated by DMP). Let $y_{t\vcentcolon t+k}$
denote a sequence of states from $y_{t}$ to $y_{t+k}$. Denote by
$y_{t\vcentcolon t+k}\vDash\phi$ if the trajectory $y_{t\vcentcolon t+k}$
satisfies a wTLTL formula $\phi$. More expressions can be achieved
by combing temporal and Boolean operators. For instance, $y_{t\vcentcolon t+k}\vDash\Square\phi$
indicates that $\phi$ is satisfied for every subtrajectory $y_{t^{'}\vcentcolon t+k}$,
$\forall t^{'}\in\left[t,t+k\right)$, $y_{t\vcentcolon t+k}\vDash\lozenge\phi$
indicates that $\phi$ is satisfied for at least one subtrajectory
$y_{t^{'}\vcentcolon t+k}$ for some $t^{'}\in\left[t,t+k\right)$,
and $y_{t\vcentcolon t+k}\vDash\phi\mathcal{T}\psi$ indicates that
$\phi$ is satisfied at least once before $\psi$ is satisfied between
$t$ and $t+k$. 

The wTLTL has both qualitative and quantitative semantics. The qualitative
semantics indicates whether or not a trajectory satisfies a specification,
while the quantitative semantics (also referred to as the robustness)
quantifies the degree of satisfaction of a specification. 
\begin{defn}[wTLTL Robustness]
\label{def:wTLTL-Robustness} Given a specification $\varphi$ and
a trajectory $y_{t\vcentcolon t+k}$, the wTLTL robustness is defined
as
\[
\begin{aligned}\rho^{w}\left(y_{t\vcentcolon t+k},\top\right) & =\rho_{max}^{w},\\
\rho^{w}\left(y_{t\vcentcolon t+k},f\left(y_{t}\right)<c\right) & =c-f\left(y_{t}\right),\\
\rho^{w}\left(y_{t\vcentcolon t+k},\lnot\phi\right) & =-\rho^{w}\left(y_{t\vcentcolon t+k},\phi\right),\\
\rho^{w}\left(y_{t\vcentcolon t+k},\land^{w}\phi_{i}\right) & =\otimes^{\land}\left(w,\left[\rho\left(y_{t\vcentcolon t+k},\phi_{1}\right),\ldots,\right.\right.\\
 & \left.\left.\rho\left(y_{t\vcentcolon t+k},\phi_{N}\right)\right]\right),\\
\rho^{w}\left(y_{t\vcentcolon t+k},\lor^{w}\psi_{i}\right) & =\oplus^{\lor}\left(w,\left[\rho\left(y_{t\vcentcolon t+k},\psi_{1}\right),\ldots,\right.\right.\\
 & \left.\left.\rho\left(y_{t\vcentcolon t+k},\psi_{N}\right)\right]\right),
\end{aligned}
\]
where $\rho_{max}^{w}\in\mathbb{R}$ represents the maximum robustness
value and $\rho\left(y_{t\vcentcolon t+k},\phi_{i}\right)$, $i=1,\ldots,N$,
are standard robustness of TLTL as defined in \cite{Li2017a}. The
aggregation functions $\otimes^{\land}$ and $\oplus^{\lor}$ are
associated with conjunctions and disjunctions, respectively, which
satisfy $\min\left(\mathbf{x}\right)\cdot\otimes^{\land}\left(w,\mathbf{x}\right)>0$
and $\max\left(\mathbf{x}\right)\cdot\oplus^{\lor}\left(w,\mathbf{x}\right)>0$,
$\forall\mathbf{x}\in\mathbb{R}^{N}$ and $\mathbf{x}\neq0$, and
are defined as
\begin{equation}
\begin{aligned}\otimes^{\land}\left(w,\mathbf{x}\right) & =\underset{i=1:N}{\min}\left\{ \left(\left(\frac{1}{2}-\bar{w_{i}}\right)\Sign\left(x_{i}\right)+\frac{1}{2}\right)\cdot x_{i}\right\} ,\\
\oplus^{\lor}\left(w,\mathbf{x}\right) & =-\otimes^{\land}\left(w,-\mathbf{x}\right),
\end{aligned}
\label{eq:aggregation function}
\end{equation}
where $\bar{w_{i}}=\frac{w_{i}}{\sum_{j=1}^{N}w_{j}}$ is the normalized
weight \cite{Mehdipour2021}. The definition of robustness for temporal
operators ($\lozenge$, $\boxempty$, $\mathcal{U}$, $\mathcal{T}$,
and $\Rightarrow$) is the same as the standard robustness of TLTL
as defined in \cite{Li2017a}.
\end{defn}
The wTLTL robustness in Def. \ref{def:wTLTL-Robustness} is \textit{sound}
in the sense that a strictly positive robustness $\rho^{w}$ indicates
satisfaction of the formula $\varphi$, and a strictly negative robustness
$\rho^{w}$ indicates violation of $\varphi$. That is, $\rho^{w}\left(y_{t\vcentcolon t+k},\varphi\right)>0$
implies $y_{t\vcentcolon t+k}\vDash\varphi$, and $\rho^{w}\left(y_{t\vcentcolon t+k},\varphi\right)<0$
implies $y_{t\vcentcolon t+k}\nvDash\varphi$. Following similar analysis
in Theorem 2 of \cite{Mehdipour2021}, it is trivial to show that
the wTLTL robustness is sound. Due to its soundness property, $\rho^{w}$
will be exploited in the subsequent development to facilitate the
design of cost functions in $\textrm{P\textrm{I}}^{\textrm{BB-TL}}$
algorithm to enable complex task manipulation with specified user
preferences.

\subsection{Smooth Approximations\label{subsec:Smooth-Approximations}}

In literature, various approaches can be employed to approximate $\textrm{min}$
and $\textrm{max}$. For instance, the max and min can be approximated
by\textcolor{black}{
\begin{equation}
\tilde{\min}\left(\left[a_{1},a_{2}\ldots,a_{m}\right]^{T}\right)=-\frac{1}{k_{1}}\log\left(\stackrel[i=1]{m}{\sum}e^{-k_{1}a_{i}}\right),\label{eq:min}
\end{equation}
}

\textcolor{black}{
\begin{equation}
\tilde{\max}\left(\left[a_{1},a_{2}\ldots,a_{m}\right]^{T}\right)=\frac{\sum_{i=1}^{m}a_{i}e^{k_{2}a_{i}}}{\sum_{i=1}^{m}e^{k_{2}a_{i}}},\label{eq:max}
\end{equation}
}where $k_{1},k_{2}>0$ are adjustable parameters \cite{Gilpin2020}.
It is shown in \cite{Gilpin2020} that (\ref{eq:min}) and (\ref{eq:max})
are under-approximation of the true minimum and maximum, i.e., it
is always true that $\tilde{\min}\left(\mathbf{a}\right)\leq\min\left(\mathbf{a}\right)$
and $\tilde{\max}\left(\mathbf{a}\right)\leq\max\left(\mathbf{a}\right)$. 

\section{Problem Formulation\label{sec:PF}}

Consider a robotic system whose motion is represented by DMPs in (\ref{eq:DMP})
with known initial position $y_{0}$ and goal position $g$, and unknown
shape parameters $\theta$. Let $y_{0:t}$ denote a finite robot trajectory
starting form $y_{0}$. The complex manipulation to be performed by
the robot is specified by a wTLTL formula $\varphi$ and its predicates
are interpreted over the trajectory $y_{0:t}$. The goal of this work
is to identify an optimal shape parameter vector of DMPs $\theta^{\ast}=\underset{\theta}{\textrm{argmin}}J,$
where $J$ is a cost function to be designed, such that the incurred
robot motion satisfies (\ref{eq:DMP}) and the formula $\varphi$,
i.e., $y_{0:t}\left(y_{0},g,\theta^{\ast}\right)\vDash\varphi.$

\begin{figure}
\centering{}\includegraphics[scale=0.24]{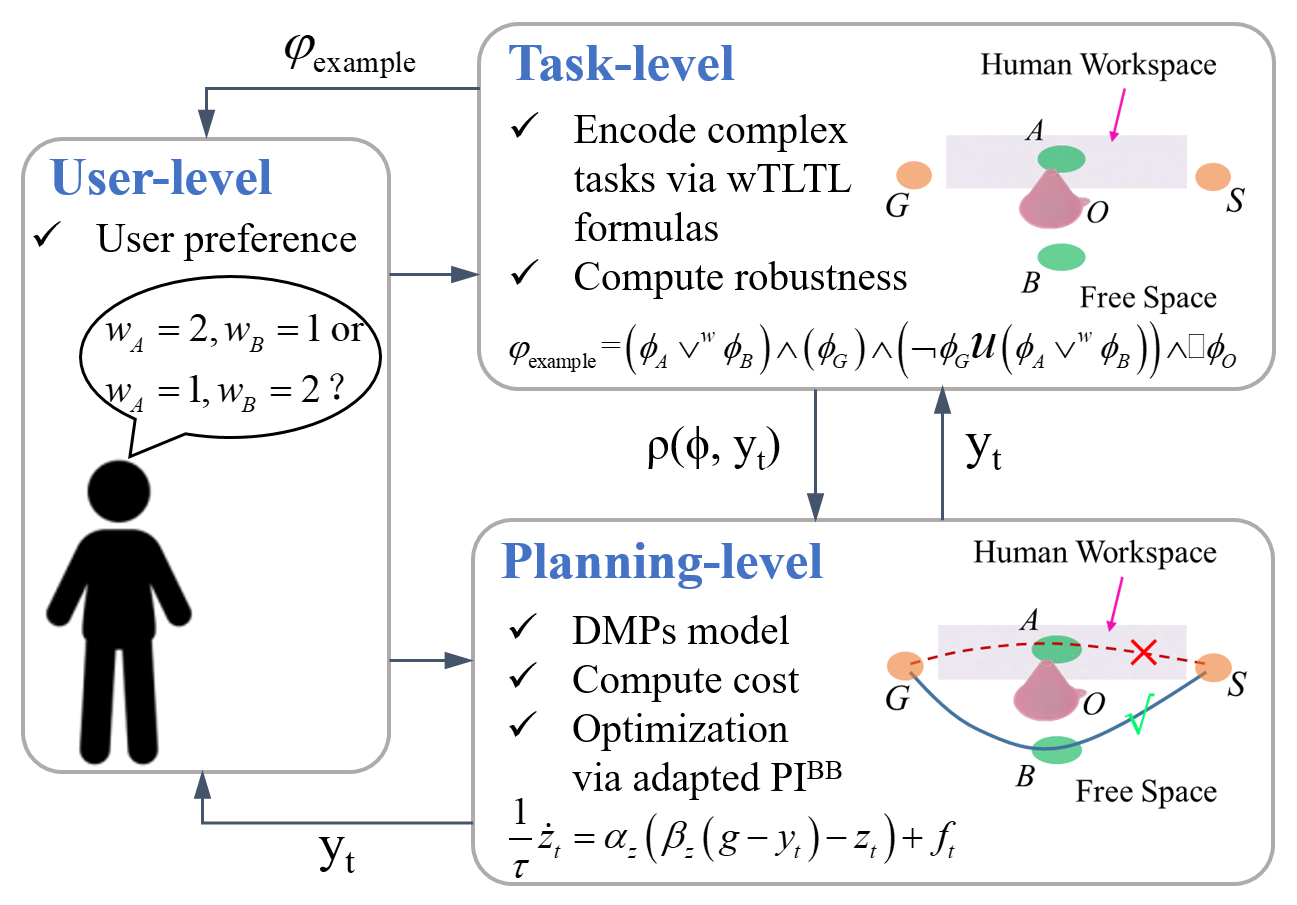}\caption{\label{fig:architecture} The architecture of $\textrm{P\textrm{I}}^{\textrm{BB-TL}}$
for complex manipulation tasks with user preferences. In the task-level,
the wTLTL specification encodes the complex manipulation task described
in Example \ref{exa:example1} and its robustness indicates how well
the task is performed. In the planning-level, satisfactory trajectories
are generated by optimizing the shape parameters of DMPs via adapted
$\textrm{P\textrm{I}}^{\textrm{BB}}$ algorithm. User preferences
are incorporated in both task and planning levels to indicate the
relative importance of subtasks or priorities of alternatives.}
\end{figure}
To address this motion planning problem, the $\textrm{P\textrm{I}}^{\textrm{BB-TL}}$
algorithm is developed. The core idea of $\textrm{P\textrm{I}}^{\textrm{BB-TL}}$
is to incorporate DMPs and wTLTL to enable complex manipulations,
where wTLTL provides interpretable task specifications to instruct
the robot what to do in the task-level, while DMPs are used to generate
robot trajectories to instruct the robot how to do in the planning-level.
User preferences are further incorporated via wTLTL and the cost function
into the task and planning level, respectively, to bias the generated
robotic trajectory towards user preferences. The architecture of $\textrm{P\textrm{I}}^{\textrm{BB-TL}}$
algorithm is illustrated in Fig. \ref{fig:architecture}. 

\begin{figure}
\centering{}\includegraphics[scale=0.24]{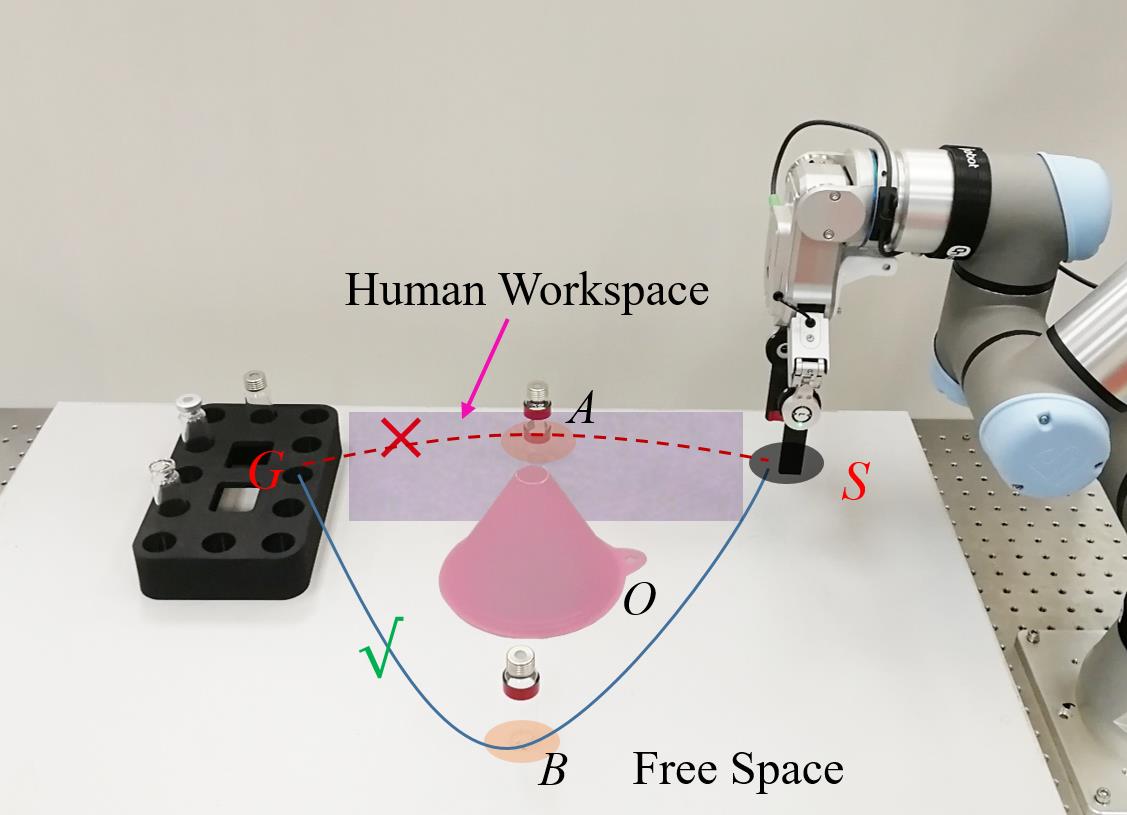}\caption{\label{fig:example} A complex manipulation task with user preference,
where $S$ represents the initial position, $G$ represents the goal
position, $O$ represents the obstacles. Human workspace is indicated
by the shaded area.}
\end{figure}

To elaborate the $\textrm{P\textrm{I}}^{\textrm{BB-TL}}$ algorithm,
a running example is used throughout this work.
\begin{example}
\label{exa:example1} Consider a task that requires the end effector
of a manipulator to transport chemical reagent from the initial position
$S$ to the reagent rack $G$ in Fig. \ref{fig:example}. The manipulator
is tasked to visit region $A$ (in human workspace) or $B$ (in free
space) before reaching $G$, and do not visit $G$ until $A$ or $B$
is visited, and always avoid obstacle $\mathit{O}$. Such a task can
be encoded by a wTLTL formula $\varphi_{example}=\left(\phi_{A}\lor^{w}\phi_{B}\right)\land\phi_{G}\land\left(\lnot\phi_{G}\mathscr{\mathcal{U}}\left(\phi_{A}\lor^{w}\phi_{B}\right)\right)\land\oblong\varphi_{O},$
where $\phi_{A}$, $\phi_{B},$ $\phi_{G}$, $\phi_{O}$ are predicates
corresponding to region $\mathit{A}$, $\mathit{B}$, $G$ and $\mathit{O}$,
respectively. It is worth pointing out that, for such a complex task
$\varphi_{example}$, conventional DMP-based approaches \cite{Bottasso2008,Stulp2012}
are no longer applicable. There are two main challenges to perform
$\varphi_{example}.$ The first challenge is how conventional DMP
can be extended to learn optimal shape parameters of motion primitives
for complex tasks at a time. Another challenge is how the optimal
shape parameters in DMP can be learned while considering user preferences.
For instance, suppose there are two feasible plans, i.e., the dashed
and solid lines in Fig. \ref{fig:example}. Even the path via $A$
to the goal $G$ is shorter, the path via $B$ is more favorable since
it avoids human workspace to reduce the potential collision of chemical
reagent with human operator. Such a preference should be encoded in
the motion planning strategy. 
\end{example}

\section{Cost Function Design and Policy Improvement \label{sec:Policy-CostFcn}}

This section first presents the cost function design in Sec. \ref{subsec:SmoothFcn},
which exploits wTLTL robustness to guide motion planning for complex
tasks. Section \ref{subsec:bbo} then presents how the black-box optimization
algorithm can be adapted to solve the synthesis problem.

\subsection{TLTL Robustness Guided Cost Function Design\label{subsec:SmoothFcn}}

Due to the soundness property of $\rho^{w}\left(y_{0:t},\varphi\right)$,
the goal is to design a cost function $J$ such that minimizing $J$
is equivalent to maximizing the satisfaction degree of the wTLTL specification
$\varphi$. Optimization methods can then be employed to identify
the optimal shape parameters $\theta^{*}$ that minimize the cost
function $J$. However, the traditional robustness only considers
satisfaction of a TLTL formula at the most extreme sub-formulas, hindering
the optimization to find feasible solutions. To address this issue,
inspired by \cite{Mehdipour2021,Gilpin2020}, we refine wTLTL robustness
by accumulating and averaging the robustness over all sub-formulas.
\begin{defn}[Smoothed wTLTL Robustness]
\label{def:Smooth-TLTL-Robustness} Given a specification $\varphi$
and a state trajectory $y_{t:t+k}$, the smoothed wTLTL robustness
$\widetilde{\rho}$ is defined as follows:
\[
\begin{aligned}\widetilde{\rho}\left(y_{t\vcentcolon t+k},\top\right) & =\widetilde{\rho}_{max}^{w},\\
\widetilde{\rho}\left(y_{t\vcentcolon t+k},f\left(y_{t}\right)<c\right) & =c-f\left(y_{t}\right),\\
\widetilde{\rho}\left(y_{t\vcentcolon t+k},\lnot\phi\right) & =-\widetilde{\rho}\left(y_{t\vcentcolon t+k},\phi\right),\\
{\color{black}{\color{red}{\color{blue}{\color{black}\widetilde{\rho}\left(y_{t\vcentcolon t+k},\Phi\land\Psi\right)}}}} & {\color{black}{\color{red}{\color{blue}{\color{black}=\tilde{\min}\left(\rho^{w}\left(y_{t\vcentcolon t+k},\Phi\right),\rho^{w}\left(y_{t\vcentcolon t+k},\Psi\right)\right),}}}}\\
{\color{blue}{\color{red}{\color{blue}{\color{black}\widetilde{\rho}\left(y_{t\vcentcolon t+k},\Phi\lor\Psi\right)}}}} & {\color{blue}{\color{red}{\color{blue}{\color{black}=\tilde{\max}\left(\rho^{w}\left(y_{t\vcentcolon t+k},\Phi\right),\rho^{w}\left(y_{t\vcentcolon t+k},\Psi\right)\right),}}}}
\end{aligned}
\]
\begin{equation}
\begin{aligned}\widetilde{\rho}\left(y_{t\vcentcolon t+k},\Square\phi\right) & =\underset{t^{'}\in\left[t,t+k\right)}{\tilde{\min}}\left(\rho^{w}\left(y_{t\vcentcolon t+k},\phi\right)\right),\\
\widetilde{\rho}\left(y_{t\vcentcolon t+k},\lozenge\phi\right) & =\underset{t^{'}\in\left[t,t+k\right)}{\tilde{\max}}\left(\rho^{w}\left(y_{t\vcentcolon t+k},\phi\right)\right),\\
{\color{red}{\color{blue}{\color{black}\widetilde{\rho}\left(y_{t\vcentcolon t+k},\phi\mathcal{U}\psi\right)}}} & {\color{red}{\color{blue}{\color{black}=\underset{t^{'}\in\left[t,t+k\right)}{\tilde{\max}}\left(\tilde{\min}\left(\rho^{w}\left(y_{t^{'}\vcentcolon t+k},\psi\right)\right),\right.}}}\\
{\color{red}} & {\color{blue}{\color{red}{\color{blue}{\color{black}\left.\underset{t^{''}\in\left[t,t^{'}\right)}{\tilde{\textrm{min}}}\left(\rho^{w}\left(y_{t^{''}\vcentcolon t^{'}},\psi\right)\right)\right),}}}}\\
\widetilde{\rho}\left(y_{t\vcentcolon t+k},\phi\mathcal{T}\psi\right) & =\underset{t^{'}\in\left[t,t+k\right)}{\tilde{\max}}\left(\tilde{\min}\left(\rho^{w}\left(y_{t^{'}\vcentcolon t+k},\psi\right)\right),\right.\\
 & \left.\underset{t^{''}\in\left[t,t^{'}\right)}{\tilde{\max}}\left(\rho^{w}\left(y_{t^{''}\vcentcolon t^{'}},\psi\right)\right)\right),\\
{\color{red}{\color{blue}{\color{black}\widetilde{\rho}\left(y_{t\vcentcolon t+k},\phi\mathcal{\Rightarrow}\psi\right)}}} & {\color{red}{\color{blue}{\color{black}=\tilde{\max}\left(-\rho^{w}\left(y_{t\vcentcolon t+k},\phi\right),\rho^{w}\left(y_{t\vcentcolon t+k},\psi\right)\right).}}}
\end{aligned}
\label{eq:smooth robustness}
\end{equation}
where $\Phi=\land^{w}\phi_{i}$ and $\Psi=\lor^{w}\psi_{i}$. 
\end{defn}
In Def. \ref{def:Smooth-TLTL-Robustness}, $\tilde{\min}$ and $\tilde{\max}$
are smoothed version of the conventional $\min$ and $\max$ operators
which are calculated by (\ref{eq:min}) and (\ref{eq:max}) as discussed
in Section \ref{subsec:Smooth-Approximations}.
\begin{thm}
\label{thm:theorem2}Given a wTLTL specification $\varphi$ and a
state trajectory $y_{t:t+k}$, for any pre-defined error bound $\epsilon>0$,
there exist $\overline{k}_{1}$ and $\overline{k}_{2}$ such that
$0\leq\rho^{w}\left(y_{t:t+k},\varphi\right)-\widetilde{\rho}\left(y_{t:t+k},\varphi\right)\leq\epsilon$
for all $k_{1}\geq\bar{k}_{1}$, $k_{2}\geq\bar{k}_{2}$, where $k_{1}$
and $k_{2}$ are tuning parameters defined in (\ref{eq:min}) and
(\ref{eq:max}).
\end{thm}
The proof of Theorem \ref{thm:theorem2} is omitted, since it is a
trivial extension of Theorem 2 in \cite{Gilpin2020}. Theorem \ref{thm:theorem2}
indicates that the smoothed wTLTL robustness $\widetilde{\rho}$ is
an under-approximation of $\rho^{w}$, and thus it provides a sufficient
condition for the satisfaction check of wTLTL specification $\varphi$.
That is, if $\widetilde{\rho}\left(y_{t:t+k},\varphi\right)>0$, it
is always true that $\rho^{w}\left(y_{t:t+k},\varphi\right)>0$, which
implies $y_{t:t+k}\vDash\varphi$. Theorem \ref{thm:theorem2} also
indicates that the smoothed wTLTL robustness $\widetilde{\rho}$ gradually
approaches the true robustness $\mathit{\rho}$ as $k_{1}$ and $k_{2}$
increase, which means that the approximation will not hide any potential
solutions if $k_{1}$ and $k_{2}$ are sufficiently large.

By Theorem \ref{thm:theorem2}, we design the cost function $J$ as
\begin{equation}
J\left(\widetilde{\rho}\left(y_{0:t},\varphi\right)\right)=\left\{ \begin{array}{cc}
-\widetilde{\rho}\left(y_{0:t},\varphi\right), & \text{ if }\widetilde{\rho}\left(y_{0:t},\varphi\right)<0,\\
0, & \text{otherwise,}
\end{array}\right.\label{eq:wTLTL cost function}
\end{equation}
which indicates that minimizing the cost function (i.e., $J\left(\widetilde{\rho}\left(y_{0:t},\varphi\right)\right)\rightarrow0$)
can lead to $\widetilde{\rho}\left(y_{0:t},\varphi\right)\geq0$,
resulting in that $y_{0:t}\vDash\varphi$ due to the soundness property
of wTLTL robustness. 

Based on (\ref{eq:wTLTL cost function}), the problem considered in
Sec. \ref{sec:PF} can then be formally formulated as 
\begin{equation}
\begin{aligned}\theta^{\ast} & =\underset{\theta}{\textrm{argmin}}J\left(\widetilde{\rho}\left(y_{0:t},\varphi\right)\right),\\
\text{s.t.} & \;(\ref{eq:DMP})\text{ are satisfied},
\end{aligned}
\label{eq:PF_new}
\end{equation}
where the constraint $y_{0:t}\left(y_{0},g,\theta^{\ast}\right)\vDash\varphi$
is implicitly encoded in the the cost function $J$ via the smoothed
wTLTL robustness. Note that the goal of (\ref{eq:PF_new}) is to identify
a satisfactory trajectory, rather than an optimal trajectory with
largest robustness, with respect to logical and temporal specifications
and user preferences. Ongoing research is to leverage the wTLTL robustness
in the design of the cost function to facilitate the identification
of the optimal trajectory. 

\subsection{TLTL Guided Black-Box Optimization\label{subsec:bbo} }

After designing the optimization problem in (\ref{eq:PF_new}), the
next step is to identify the optimal shape parameters $\theta^{*}$
with respect to $J$. Two widely used approaches to perform this optimization
are reinforcement learning (RL) and black-box optimization (BBO).
As discussed in \cite{Chatzilygeroudis2020} and \cite{Stulp2012a},
$\textrm{P\textrm{I}}^{\textrm{2}}$ is a RL algorithm which uses
reward information during exploratory policy executions, while $\textrm{P\textrm{I}}^{\textrm{BB}}$,
a variant of $\textrm{P\textrm{I}}^{\textrm{2}}$ with constant exploration
and without temporal averaging, is a BBO algorithm that uses the total
reward during execution, which enables that the utility function can
be treated as a black box. Despite the lack of theoretical guarantees,
strong empirical evidence shows that $\textrm{P\textrm{I}}^{\textrm{BB}}$
has equal or superior performance than $\textrm{P\textrm{I}}^{\textrm{2}}$
in terms of convergence speed and robustness of policy improvement.
In addition, as a special case of covariance matrix adaptation evolutionary
strategy, $\textrm{P\textrm{I}}^{\textrm{BB}}$ is a global optimization
method\cite{Stulp2012a,Hansen2001}. 

\begin{algorithm}
\caption{\label{Alg:PIBB}wTLTL Guided $\textrm{P\textrm{I}}^{\textrm{BB-TL}}$
Algorithm for DMPs}

\scriptsize

\singlespacing

\begin{algorithmic}[1]

\Procedure {Input: } {the initial shape parameter $\theta^{init}$;
the wTLTL specification $\varphi$; the exploration levels $\lambda^{init}$,
$\lambda^{min}$, and $\lambda^{max}$; the sample size $M$; the
eliteness parameter $h$}

{Output: } {the optimal shape parameters $\theta^{*}$ }

{Initialization: } {set $\theta=\theta^{init}$ , $\Sigma=\lambda^{init}I$
}

\While {cost $J\left(\widetilde{\rho}\left(y_{0:t},\varphi\right)\right)$
not converged}

\State Exploration: sample parameters and compute costs

\For { each $m\in M$ }

\State $\theta_{m}\sim\mathcal{N}\left(\theta,\Sigma\right)$

\State Generate a trajectory $y_{0:t}^{m}$ based on current $\theta_{m}$
according to (\ref{eq:DMP})

\State Compute the $\widetilde{\rho}\left(y_{0:t}^{m},\varphi\right)$
according to Def. \ref{def:wTLTL-Robustness} and Def. \ref{def:Smooth-TLTL-Robustness}

\State Compute the cost $J_{m}$ according to (\ref{eq:wTLTL cost function})

\EndFor

\State $J^{min}=\textrm{min}\left\{ J_{m}\right\} _{m=1}^{M}$

\State $J^{max}=\textrm{max}\left\{ J_{m}\right\} _{m=1}^{M}$

\State Evaluation: compute weight for each sample

\For { each $m\in M$ }

\State $P_{m}=\frac{exp\left(-h\frac{J_{m}-J^{min}}{J^{max}-J^{min}}\right)}{\sum_{l=1}^{M}exp\left(-h\frac{J_{l}-J^{min}}{J^{max}-J^{min}}\right)}$

\EndFor

\State Update: weighted averaging over $M$ samples

\State $\Sigma\leftarrow\sum_{m=1}^{M}\left[P_{m}\left(\theta_{m}-\theta\right)\left(\theta_{m}-\theta\right)^{T}\right]$

\State $\Sigma^{new}\leftarrow\textrm{boundcovar}\left(\Sigma,\lambda^{min},\lambda^{max}\right)$

\State $\theta^{new}\leftarrow\stackrel[m=1]{M}{\sum}\left[P_{m}\theta_{m}\right]$

\EndWhile

\EndProcedure

\end{algorithmic}
\end{algorithm}

Motivated by the discussion above, this section presents how the stochastic
optimization algorithm $\textrm{P\textrm{I}}^{\textrm{BB}}$ can be
adapted in our $\textrm{P\textrm{I}}^{\textrm{BB-TL}}$ to solve the
optimization problem\footnote{In contrast to the traditional TLTL robustness in \cite{Li2017a},
the smoothed wTLTL robustness $\widetilde{\rho}$ is differentiable,
which makes it also suitable for gradient-based optimization methods,
e.g., RL-based optimization methods \cite{Gilpin2020}. Since BBO
based optimization in general outperforms RL based optimization, this
work focuses on adapted stochastic optimization algorithm $\textrm{P\textrm{I}}^{\textrm{BB}}$.} in (\ref{eq:PF_new}). Specifically, the $\textrm{P\textrm{I}}^{\textrm{BB-TL}}$
algorithm is outlined in Alg. \ref{Alg:PIBB}, and the shape parameters
are updated following the subsequent steps: 
\begin{enumerate}
\item Initialization (line 1): Set the mean and covariance $\left\langle \theta,\Sigma\right\rangle $
to $\left\langle \theta^{init},\lambda^{init}I\right\rangle $, where
$I$ is an identity matrix with appropriate dimension. 
\item Exploration (line 3-9): Sample $M$ parameter vectors $\theta_{m}$,
$m=1,\ldots M$, from $\mathcal{N}\left(\theta,\Sigma\right)$ according
to $\left\{ \theta_{m}=\theta+\epsilon_{m}\right\} _{m=1}^{M}$, where
$\epsilon_{m}$ is sampled from a zero-mean Gaussian distribution
with variance $\Sigma$. Compute the cost $J_{m}=J\left(\widetilde{\rho}\left(y_{0:t},\varphi\right)\right)$
of each sample.
\item Evaluation (line 10-15): Compute the weight $P\in\mathbb{R}^{M}$
for the $M$ samples, where the $m$th entry $P_{m}\in\left[0,1\right]$
and $\sum_{m=1}^{M}P_{m}=1$. Specifically, using the normalized exponentiation
function, $P_{m}$ is computed as 
\begin{equation}
\begin{array}{c}
P_{m}=\frac{\exp\left(-h\frac{J_{m}-J^{min}}{J^{max}-J^{min}}\right)}{\sum_{r=1}^{M}\exp\left(-h\frac{J_{r}-J^{min}}{J^{max}-J^{min}}\right)},\end{array}
\end{equation}
where $J^{min}=\textrm{min}\left\{ J_{m}\right\} _{m=1}^{M}$, $J^{max}=\textrm{max}\left\{ J_{m}\right\} _{m=1}^{M}$,
and $h$ is the eliteness parameter. If a large $h$ is used, only
a few samples will contribute to the weighted averaging. If $h=0$,
no learning would occur since all given samples have equal weight
independent of the cost. In general, low-cost samples have higher
weights, and vice versa.
\item Update (line 16-19): Update the parameters $\left\langle \theta,\Sigma\right\rangle $
to $\left\langle \theta^{new},\Sigma^{new}\right\rangle $ with weighted
averaging according to
\begin{equation}
\Sigma^{new}\leftarrow\textrm{boundcovar}\left(\Sigma,\lambda^{min},\lambda^{max}\right)\label{eq:update covariance}
\end{equation}
and 
\begin{equation}
\theta^{new}=\stackrel[m=1]{M}{\sum}\left[P_{m}\theta_{m}\right],\label{eq:update mean}
\end{equation}
where 
\begin{equation}
\Sigma=\stackrel[m=1]{M}{\sum}\left[P_{m}\left(\theta_{m}-\theta\right)\left(\theta_{m}-\theta\right)^{T}\right]\label{eq:sigma}
\end{equation}
and $\textrm{boundcovar}\left(\Sigma,\lambda^{min},\lambda^{max}\right)$
is a function that restricts the eigenvalues $\lambda$ of the covariance
matrix $\Sigma$ within $\left[\lambda^{min},\lambda^{max}\right]$,
i.e., if $\lambda<\lambda^{min}$, then $\lambda=\lambda^{min}$;
if $\lambda>\lambda^{max}$, then $\lambda=\lambda^{max}$. The low-cost
samples contribute more to the update since they have higher weights,
leading to that $\theta$ moves towards its optimum $\theta^{\ast}$.
\end{enumerate}
\begin{figure}
\centering{}\includegraphics[scale=0.19]{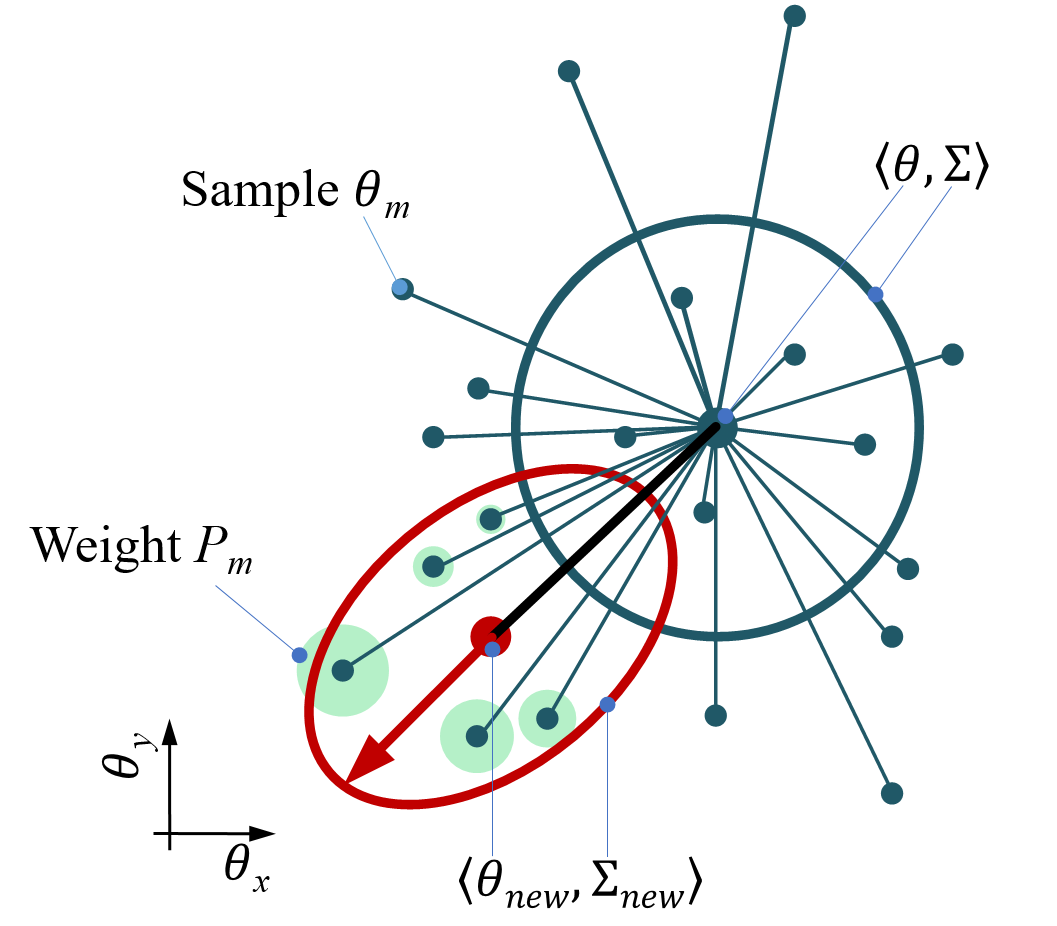}\caption{\textcolor{red}{\label{fig:visualization of update}}Visualization
of one parameter update using $\textrm{P\textrm{I}}^{\textrm{BB-TL}}$
in a 2-D search space. Sampling $M$ parameter vectors when the current
parameter $\theta$ has not converged to the optimum $\theta^{*}$.
The samples closer to $\theta^{*}$ have larger weights (visualized
as green circles) than those further away from $\theta^{*}$. Since
the new parameters are weighted, the mean $\theta$ moves in the direction
towards the high-weight and low-cost samples during the update, so
that $\mathit{\theta}$ moves closer to $\theta^{*}$.}
\end{figure}
The visualization of one parameter update using $\textrm{P\textrm{I}}^{\textrm{BB-TL}}$
in a 2-D search space is shown in Fig. \ref{fig:visualization of update}.
We sample $M$ parameter vectors when the current parameter $\theta$
has not converged to the optimum $\theta^{*}$. The samples closer
to $\theta^{*}$ have larger weights (visualized as green circles)
than those further away from $\theta^{*}$. Since the new parameters
are weighted, the mean $\theta$ moves in the direction towards the
high-weight and low-cost samples during the update, so that $\mathit{\theta}$
moves closer to $\theta^{*}$. The same idea applies to the covariance
matrix. When the parameters update, its maximum eigenvalue $\mathit{\lambda}$
of the covariance matrix increases, which results in an elongated
covariance matrix such that the eigenvector points towards the direction
of $\theta^{*}$ (visualized as an arrow). Following the steps above
until $J_{m}\rightarrow0$, the optimal $\theta^{\ast}$ can be obtained,
which in turn yields a trajectory $y_{0:t}$ that satisfies the wTLTL
specifications $\varphi$. Given that the sample size is $M$ and
the number of updates required for cost convergence is $\varLambda$,
the computational complexity of $\textrm{P\textrm{I}}^{\textrm{BB-TL}}$
is $O(\varLambda M)$ according to Alg. \ref{Alg:PIBB}.

\section{Case Studies\label{sec:Case}}

In this section, the developed temporal logic guided $\textrm{P\textrm{I}}^{\textrm{BB-TL}}$
algorithm is implemented in Python 3.5 on Ubuntu 16.04. To validate
the effectiveness of our approach, we first carry out two simulations
on a Mac with 3.40 GHz Intel Xeon E-2224 CPU and 16 GB RAM, and then
validate this approach in a real-world case using Universal Robot
UR5e. 

\subsection{Simulation Results}

\begin{figure}
\centering{}\includegraphics[scale=0.19]{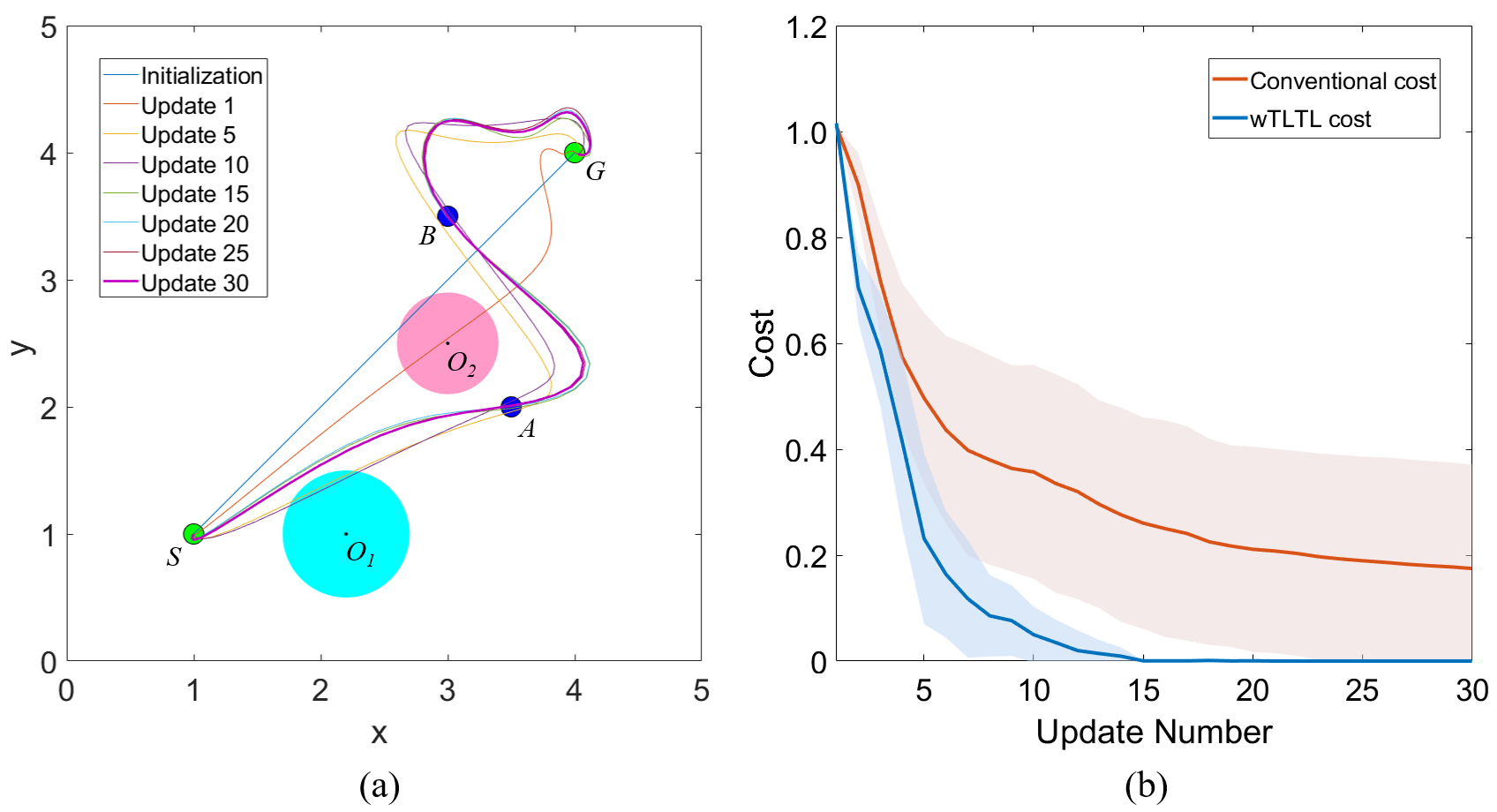}\caption{\label{fig:case_study1} (a) The evolution of generated trajectories
using $\textrm{P\textrm{I}}^{\textrm{BB-TL}}$ algorithm in Case 1
for a sequential goal reaching task with obstacle avoidance. (b) Learning
curves of the wTLTL guided cost and the conventional cost trained
with $\textrm{P\textrm{I}}^{\textrm{BB}}$ in Case 1, respectively.
The solid line shows the mean of the cost and the shaded area represents
the corresponding standard deviation of 20 training results.}
\end{figure}

Consider a 2-D system whose motion is described by the DMPs in (\ref{eq:DMP}).
The developed wTLTL guided $\textrm{P\textrm{I}}^{\textrm{BB-TL}}$
algorithm is implemented to optimize the shape parameters $\theta$
of DMPs. The parameters of $\textrm{P\textrm{I}}^{\textrm{BB-TL}}$
are set as follows. The initial shape parameters vector $\theta^{init}$
is a zero vector. The initial, minimum, and maximum exploration are
set to $\lambda^{init}=0.05$, $\lambda^{min}=0.05$, $\lambda^{max}=None$,
respectively. The number of trials per update is set to $\Gamma=20$,
the number of basis functions for 1-D DMP $\kappa=10$, the eliteness
parameter is set to $h=10$. 

\textbf{Case 1: }In this case, to show the capability of generating
trajectories for complex manipulations, the conventional cost function
from \cite{Stulp2012} is treated as a baseline and compared with
the developed wTLTL guided cost function. Specifically, we consider
a sequential goal reaching task, in which, as shown in Fig. \ref{fig:case_study1},
the robot is required to go from $S$ to $G$, and sequentially visit
$A$ and $B$ while avoiding obstacles $O_{1}$ and $O_{2}$. Such
a complex task can be formulated using wTLTL specification as
\begin{equation}
\varphi_{case1}=\left(\phi_{A}\mathcal{T}\phi_{B}\right)\land\left(\lnot\phi_{B}\mathcal{U}\phi_{A}\right)\land\oblong\varphi_{O},\label{eq:case1}
\end{equation}
where $\varphi_{O}=\wedge_{i=1,2}d_{O_{i}}>r_{O_{i}}$ with $d_{O_{i}}$
representing the Euclidean distance between the robot and the obstacle
$O_{i}$ and $r_{O_{i}}$ representing the radius of obstacle $O_{i}$.
In English, $\varphi_{case1}$ means ``visit $A$ then $B$, do not
visit $B$ until $A$ is visited, and always avoid the obstacles $O_{1}$
and $O_{2}$''. 

Since conventional cost function cannot deal with user preferences,
to be a fair comparison, the user preferences in wTLTL guided cost
function are not considered in this case, i.e., we set all weights
$w_{i}=1,$ $\forall i\in\left\{ 1,\ldots,N\right\} $ in implementation.
The smoothed wTLTL robustness and cost function of $\textrm{P\textrm{I}}^{\textrm{BB-TL}}$
can be obtained by following (\ref{eq:smooth robustness}) and (\ref{eq:wTLTL cost function}),
respectively. Following the design in \cite{Stulp2012}, the conventional
cost function is adapted to the particular task $\varphi_{case1}$
as
\begin{equation}
\bar{J}=-c_{1}J_{O}+c_{2}J_{G}\label{eq:J_bar}
\end{equation}
where $c_{1},c_{2}\in\mathbb{R}^{+}$ are relative weights, i.e. $c_{1}=0.6$,
$c_{2}=0.5$. In (\ref{eq:J_bar}), $J_{O}=\sum_{i=1,2}(d_{O_{i}}-r_{O_{i}})$
where $d_{O_{i}}-r_{O_{i}}=0$ if $d_{O_{i}}>r_{O_{i}}$, and $J_{G}=d_{A}+d_{B}$
where $d_{A}$ and $d_{B}$ are the Euclidean distance between the
robot and the region $A$ and $B$, respectively. 

The average learning time for the shape parameters $\theta\in\mathit{\mathbb{R}^{\textrm{10}}}$
of DMPs with respect to the conventional cost function and the wTLTL
guided cost function are $7.89$s and $6.94$s, respectively. Fig.
\ref{fig:case_study1} (a) shows the generated trajectories after
updates using $\textrm{P\textrm{I}}^{\textrm{BB-TL}}$, which indicates
$\varphi_{case1}$ is successfully completed. The evolution of the
costs are shown in Fig. \ref{fig:case_study1} (b), which indicates
that the wTLTL guided cost function outperforms the conventional cost
function in the sense that the cost converges faster. 

\begin{figure}
\centering{}\includegraphics[scale=0.19]{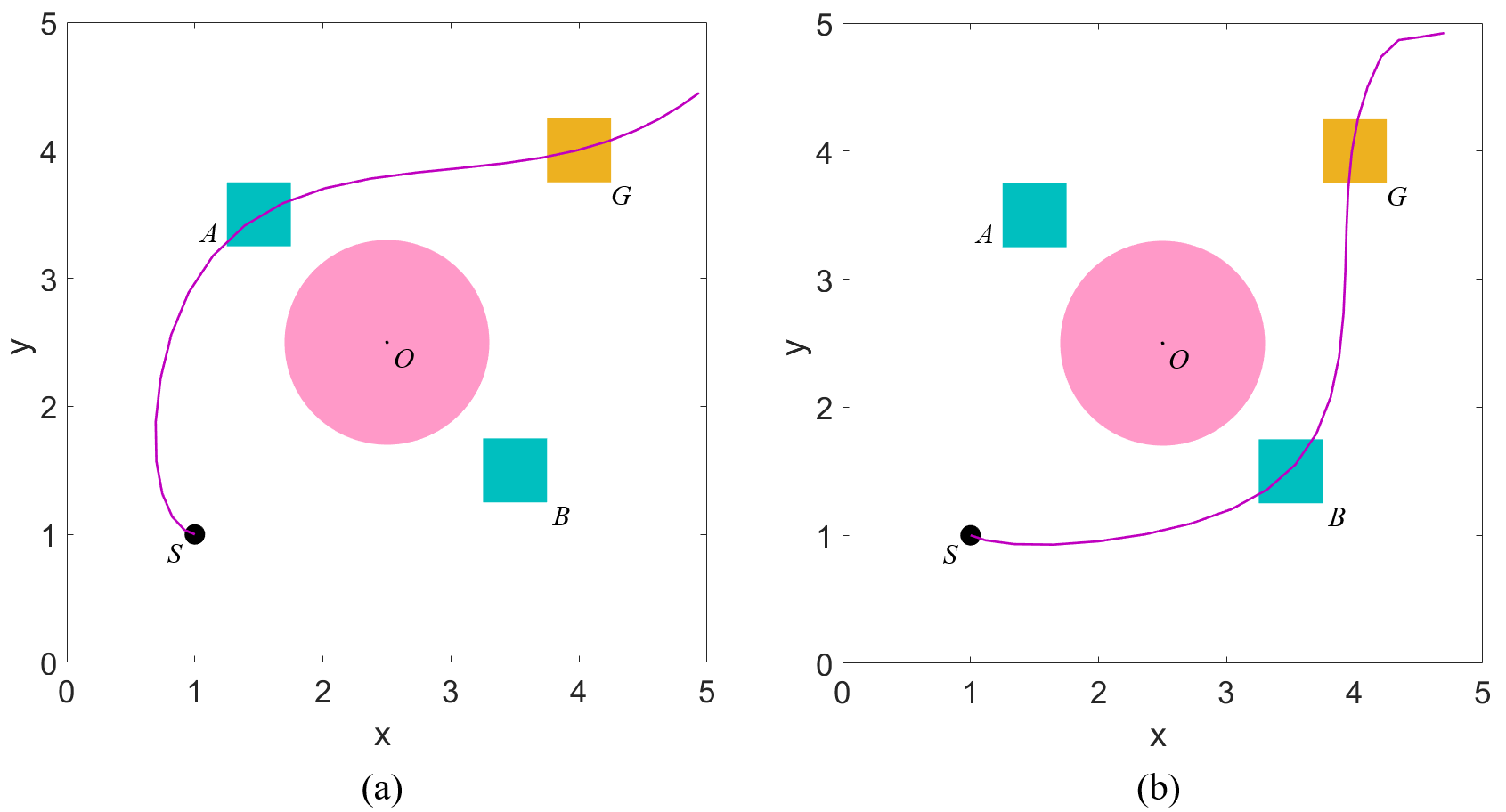}\caption{\label{fig:case_study2} Simulated trajectories of $\varphi_{case2}$
with $w_{A}=2,w_{B}=1$ in (a) and $w_{A}=1,w_{B}=2$ in (b).}
\end{figure}

\textbf{Case 2: }To show the capability of handling user preferences
in $\textrm{P\textrm{I}}^{\textrm{BB-TL}}$, we consider a similar
case in \cite{Mehdipour2021}. This case considers a task that requires
the robot to visit region $A$ or $B$, then visit region $G$, and
do not visit $G$ until either $A$ or $B$ is visited, while always
avoiding obstacle $O$, as shown in in Fig. \ref{fig:case_study2}.
Such a task can be written in wTLTL formula as
\begin{equation}
\varphi_{case2}=\left(\phi_{A}\lor^{w}\phi_{B}\right)\land\phi_{G}\land\left(\lnot\phi_{G}\mathscr{\mathcal{U}}\left(\phi_{A}\lor^{w}\phi_{B}\right)\right)\land\oblong\varphi_{O},\label{eq:case2}
\end{equation}
where $\varphi_{O}=d_{O}>r_{O}$, $d_{O}$ is the Euclidean distance
between the trajectory and obstacle $O$, $r_{O}$ is the radius of
obstacle $O_{i}$. 

The task $\varphi_{case2}$ indicates that the goal $G$ can be reached
by a path that traverses either region $A$ or $B$. To reflect the
user preferences, $w_{A}$ and $w_{B}$ are used to indicate the weights
about which path is more preferred, i.e., region $A$ is preferred
if $w_{A}>w_{B}$ and region $B$ is preferred otherwise. The weights
are normalized by $\bar{w_{i}}=\frac{w_{i}}{\sum_{j=1}^{N}w_{j}}$.
The wTLTL robustness and the cost function are then obtained following
the methods in Section \ref{sec:Policy-CostFcn}, and then optimized
using $\textrm{P\textrm{I}}^{\textrm{BB-TL}}$. The generated trajectories
are shown in Fig. \ref{fig:case_study2} (a) and (b), and the learning
times are $7.97\textrm{s}$ and $8.04\textrm{s}$, respectively. It
is clear from Fig. \ref{fig:case_study2} that user preferences can
be incorporated in trajectory generation, i.e., regions with higher
priority (importance) are likely to be visited if higher disjunction
(conjunction) aggregator weights are assigned accordingly. 

\subsection{Experimental Results}

\begin{figure*}
\centering{}\includegraphics[scale=0.45]{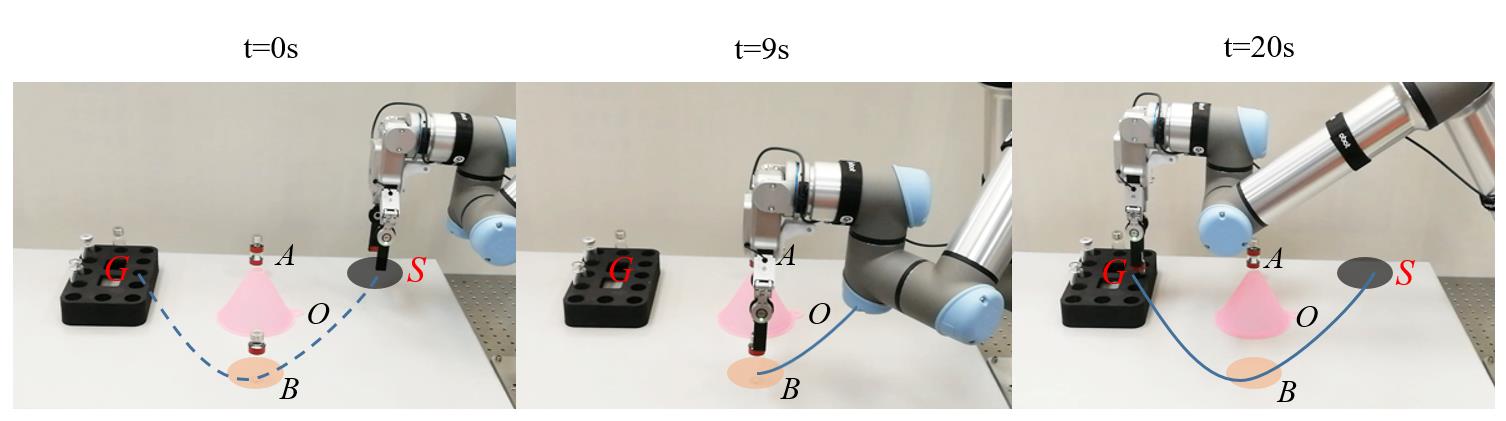}\caption{\label{fig:process of multi-task} Snapshots of a manipulation task
with complex logic and temporal constraints. The end effector of UR5e
transports chemical reagent from $S$ to the reagent rack $\mathit{G}$
via region $B$ with user preference. }
\end{figure*}
 To evaluate practical performance, the developed $\textrm{P\textrm{I}}^{\textrm{BB-TL}}$
is validated on a physical UR5e robot in a 3D workspace, which includes
2D trajectory planning and following and 3D grasp/release actions.
The 2D trajectory is generated offline using $\textrm{P\textrm{I}}^{\textrm{BB-TL}}$
with user preferences and temporal logic constraints, while the grasp/release
actions are performed in the 3D workspace following embedded library
functions. The experimental layout is shown in Example \ref{exa:example1}
and the task to be performed by the manipulator is the same as $\varphi_{case2}$
in (\ref{eq:case2}). Following Section \ref{sec:Policy-CostFcn},
the wTLTL robustness and the cost function with user preference can
be obtained. By setting a larger weight, a trajectory that passes
region $\mathit{B}$ can be trained via Alg. \ref{Alg:PIBB} offline.
Such a path reflects the user's preference on safety, i.e., prefer
to avoid human workspace when carrying chemical reagent, but with
the cost of traveling a longer path to reach the goal $G$. Fig. \ref{fig:process of multi-task}
shows the snapshots of the motion of the end effector of the UR5e
manipulator, which indicates that the end effector first goes to region
$\mathit{B}$ from $\mathit{S}$ to grasp the chemical reagent with
user preference and then transports it to the reagent rack $\mathit{G}$
while avoiding obstacle $\mathit{O}$. The experiment video is provided.\footnote{https://www.youtube.com/watch?v=IGnVKqC-T-A}

\subsection{Discussions }

As indicated by the simulation results, the wTLTL guided cost function
outperforms the conventional cost function, mainly due to the incorporation
of temporal logic in the design of DMPs. Since the sequential goal
reaching task can also be solved by the $\textrm{P\textrm{I}}^{\textrm{2}}$SEQ
algorithm in \cite{Stulp2012}, the performance of $\textrm{P\textrm{I}}^{\textrm{2}}$SEQ
and $\textrm{P\textrm{I}}^{\textrm{BB-TL}}$ are compared. Table \ref{tab:compare}
shows, under the same conditions, the number of DMPs and the number
of optimization parameters required in Case 1, Case 2 and UR5e experiment
for $\textrm{P\textrm{I}}^{\textrm{2}}$SEQ and $\textrm{P\textrm{I}}^{\textrm{BB-TL}}$,
respectively. Given a $n$-goal-reaching task in a $\eta$-D space
with $\kappa$ basis functions, $\eta n$ DMPs with $\eta n\kappa$
weight parameters need to be trained using $\textrm{P\textrm{I}}^{\textrm{2}}$SEQ
while only $\eta$ DMPs with $\eta\kappa$ weight parameters are needed
using $\textrm{P\textrm{I}}^{\textrm{BB-TL}}$ to generate a satisfactory
trajectory. Therefore, $\textrm{P\textrm{I}}^{\textrm{BB-TL}}$ is
more efficient in the sense that less parameters need to be trained.
In addition, due to the resemblance to human natural language, the
employment of temporal logic significantly increases the interpretability
of the cost function in DMPs. That is, complex tasks can be conveniently
formulated in temporal logic specifications and then encoded in the
cost function of DMPs to facilitate the generation of satisfactory
trajectories. Such an advantage is not available in the conventional
DMP-based methods. Moreover, by using wTLTL, our approach can further
incorporate user preferences in the trajectory generation, which has
less been considered in many existing DMP-based methods. It is worth
mentioning that, as a special case of covariance matrix adaptation
evolutionary strategy, $\textrm{P\textrm{I}}^{\textrm{BB-TL}}$ is
a global optimization method \cite{Stulp2012a,Hansen2001} while $\textrm{P\textrm{I}}^{\textrm{2}}$SEQ
is a gradient-based method and may converge to a local optimum.

\begin{table}
\caption{\label{tab:compare}\textcolor{red}{{} }Comparison of the number of
DMPs and the number of optimization parameters using $\textrm{P\textrm{I}}^{\textrm{2}}$SEQ
and $\textrm{P\textrm{I}}^{\textrm{BB-TL}}$}
\centering{}\resizebox{0.47\textwidth}{!}{%%
\begin{tabular}{c|c>{\centering}p{1.2cm}>{\centering}p{1.2cm}|c>{\centering}p{1.2cm}>{\centering}p{1.2cm}}
\hline 
 & \multicolumn{3}{c|}{Number of DMPs} & \multicolumn{3}{c}{Number of Parameters}\tabularnewline
\hline 
 & Case 1 & Case 2 & UR5e & Case 1 & Case 2 & UR5e\tabularnewline
$\textrm{P\textrm{I}}^{\textrm{2}}$SEQ & 6 & 4 & 4 & 60 & 40 & 40\tabularnewline
$\textrm{P\textrm{I}}^{\textrm{BB-TL}}$ & 2 & 2 & 2 & 20 & 20 & 20\tabularnewline
\hline 
\end{tabular}}
\end{table}

\section{Conclusion }

A temporal logic guided $\textrm{P\textrm{I}}^{\textrm{BB-TL}}$ algorithm
is developed in this work to generate desired motions for complex
manipulation tasks with user preferences. The integration of wTLTL
not only enables the encoding of complex tasks that involve a sequence
of logically organized action plans with user preferences, but also
provides a convenient and efficient means to design the cost function.
The black-box optimization is adapted to identify optimal shape parameters
of DMPs to enable motion planning of robotic systems. Simulation and
experiment demonstrate its success in handling complex manipulations;
however, current research mainly focus on motion generation in static
environments. Ongoing research will consider online adaptive optimization
and reactive temporal logic planning to extend $\textrm{P\textrm{I}}^{\textrm{BB-TL}}$
to dynamic workspace with mobile obstacles and time-varying missions. 

\bibliographystyle{IEEEtran}
\bibliography{Bibfile}

\end{document}